\documentclass[conference]{article}

\usepackage[utf8]{inputenc}
\usepackage[T1]{fontenc}
\usepackage[USenglish]{babel}
\usepackage[final]{graphicx}
        \graphicspath{{figures/}}
        \DeclareGraphicsExtensions{.pdf,.eps,.png,.jpg}
\usepackage[detect-weight]{siunitx}
\usepackage{booktabs}

\graphicspath{{figures/}}

\usepackage{tikz}
    \usetikzlibrary{arrows.meta, calc, backgrounds, decorations.pathmorphing, matrix, positioning, shapes}
    \usepackage{colors-accessible}
\usepackage[nolist,smaller]{eee-acronyms}
\usepackage{acronym-patch}
\usepackage{acronym-alwayslong}
\usepackage{amsmath}
\usepackage[noabbrev,capitalise]{cleveref}

\usepackage{enumitem}

\usepackage{suffix}
\newcommand*\gatehelper[1]{\textsc{#1}}
\newcommand*\gate[1]{\gatehelper{#1} gate}
\newcommand*\gates[1]{\gatehelper{#1} gates}
\WithSuffix{\newcommand*}\gate*[1]{\gatehelper{#1}}

\acrodef{AXI}{AXI}
\acronymalwaysshort{ASIC}
\acronymalwaysshort{AXI}
\acronymalwaysshort{FPGA}
\acronymalwaysshort{IO}
\acronymalwaysshort{RTL}

\newcommand*{\scaleplot}{0.45}

\title{An FPGA Architecture for Online Learning\\using the Tsetlin Machine}
\author{Samuel Prescott$^1$, Adrian Wheeldon$^2$, Rishad Shafik$^3$, Tousif Rahman$^3$,\\Alex Yakovlev$^3$ \& Ole-Christoffer Granmo$^4$\\\\
$^1$ University College London, UK\quad
$^2$ The University of Edinburgh, UK\\
$^3$Newcastle University, UK\quad
$^4$ The University of Agder, Norway}
\date{\today}

\begin{document}
    \maketitle
    


    \begin{abstract}
There is a need for \ac{ML} models to evolve in unsupervised circumstances. New classifications may be introduced, unexpected faults may occur, or the initial dataset may be small compared to the data-points presented to the system during normal operation. Implementing such a system using neural networks involves significant mathematical complexity, which is a major issue in power-critical edge applications.
This paper proposes a novel \ac{FPGA} infrastructure for online learning, implementing a low-complexity \ac{ML} algorithm called the \ac{TM}. This infrastructure features a custom-designed architecture for run-time learning management, providing on-chip offline and online learning. Using this architecture, training can be carried out on-demand on the \ac{FPGA} with pre-classified data before inference takes place. Additionally, our architecture provisions online learning, where training can be interleaved with inference during operation. \ac{TM} training naturally descends to an optimum, with training also linked to a threshold hyper-parameter which is used to reduce the probability of issuing feedback as the \ac{TM} becomes trained further. The proposed architecture is modular, allowing the data input source to be easily changed, whilst inbuilt cross-validation infrastructure allows for reliable and representative results during system testing. We present use cases for online learning using the proposed infrastructure and demonstrate the energy\slash performance\slash accuracy trade-offs.
\end{abstract}
    \acresetall
    
    \section{Introduction}\label{sec:intro}

Traditional \ac{ML} models are trained before use with an initial set of pre-classified data, so called \emph{batch learning}.
A recent concept has been to develop \ac{ML} models more akin to human learning, where the model learns over time and adapts to changes~\cite{london1999empowered}. This concept is called \emph{continual or online learning}~\cite{fontenla2013online}. It involves training using data as it arrives at the system. In batch learning systems this data would only be used for inference, but online learning evolves the training process under various real-world circumstances~\cite{sahoo2017online}. These may include
introducing new classes after deployment of the system~\cite{yu2021online},
mitigating faults during runtime~\cite{napolitano1999complete}, and learning new information from datapoints that were not available at design time~\cite{biswas2017machine}.

However, enabling online learning is challenging for the following two major reasons. Firstly, the learning system requires architectural provisions for managing (i.e. storing and retrieving) a large volume of data at real-time~\cite{sahoo2017online,maheshwari2023redress}. Secondly, the online training process aimed at high learning accuracy is typically characterized by iterative arithmetic routines, such as those in neural networks, together with the associated hyperparameter optimisations~\cite{wang2021enabling}. When implemented in software, these can dramatically increase the performance and energy costs -- which can render online learning infeasible~\cite{sahoo2017online}. As such, there is growing demand of accelerated hardware architectures, implemented in \ac{ASIC} or \ac{FPGA}.


The \ac{TM} is a new \ac{ML} algorithm based around propositional logic and reinforcement learning~\cite{granmo2018tsetlin}. With parallel logic based structures and low-complexity learning controls by only 2 hyperparameters, the \ac{TM} algorithm has recently shown competitive classification accuracy and energy consumption across a wide range of applications when compared with neural networks~\cite{wheeldon2020learning,bhattarai2020measuring,saha2021relational,lei2020arithmetic,wheeldon2020lowlatency} (see \cref{sec:tm-intro} for more details on the \ac{TM}). In this paper, we  aim to leverage these properties for a low-energy online learning infrastructure. Specifically, we present \iac{FPGA} architecture that forms part of the \ac{TM} ecosystem to address the following three goals: i) acceleration of \ac{TM} inference with the capability of training online; ii) acceleration of online hyper-parameter search for faster prototyping and cross-validation of datasets; and iii) as \iac{RTL} prototyping platform for \ac{ASIC}-oriented designs based on the \ac{TM}.


\subsection*{Contributions:}
This paper presents a novel, online learning accelerator architecture implemented using \ac{TM}. Our aim is to achieve at-speed learning capability, while also maintaining low energy costs. \emph{The main contributions of the paper are:}
\begin{itemize}
    \item an \ac{FPGA} tailored architectural design for an online learning system, featuring both data and learning management subsystems;
    \item a detailed analysis of supervised online training in the \ac{TM}; and
    \item experimental validations and demonstrations of the architecture using a number of use cases: i) introduction of new training data including operation in supervised mode; ii) introduction of a new class during runtime; and iii) mitigation of faults in \acp{TA} and clause outputs.
\end{itemize}


The remainder of this paper is organized as follows. Section II briefly introduces \ac{TM}. Section III and IV detail the proposed online learning architecture and integrated system operation. Section V discusses experimental results, presenting 3 different use cases. Section VI highlights some key performance tradeoffs, while Section VII concludes the paper.
    \section{The Tsetlin Machine}\label{sec:tm-intro}
\begin{figure}
        \centering
        \begin{tikzpicture}[x=1cm, y=0.5cm, node distance=0.5cm,
    >=stealth,
    thick,
    font=\sffamily\scriptsize,
    block/.style={rectangle, draw,
      text width=7em, text centered, rounded corners,
      minimum height=3ex,
    },
    fault/.style={%
	    draw=red, 
	    decorate,
	    decoration={%
		    snake,
		    segment length=2mm,
		    pre length=2mm,
		    post length=1mm,
	    },
    },
    fault block/.style={%
	    block,
	    text width=,
	    draw=black,
	    fill=qual-light2,
    },
]
	\node [block, minimum width=3em] (clause) {Clauses};
        \node [right=1.5 of clause.north east, anchor=north west, text depth=1.5em,
		block, minimum width=3em, fill=white] (ta) {Automaton Teams};
        \node [block, text width=, above left=0.2 of ta.south] {\acs{TA}};
        \node [block, text width=, above right=0.2 of ta.south] {\acs{TA}};
        \node [block, below=4 of clause,
                isosceles triangle,
                shape border rotate=-90,
                isosceles triangle apex angle=120,
                inner sep=0pt,
                minimum width=8em,
                ] (vote) {};
        \coordinate (e) at (vote.north east); \coordinate (w) at (vote.north west); \coordinate (s) at (vote.south);
        \node at ([yshift=0.5mm]barycentric cs:e=1,w=1,s=1) {Majority Voting};
        \node [block, below=of vote, text width=]
                (threshold) {Argmax};

        \draw [->] (clause) -- (vote) node [pos=.7, right] {Clause Results};
        \draw [<-] (clause.east) -- (ta.west|-clause.east);
        \path (ta.north west) -- (clause.north east) coordinate [midway] (incmid);
        \node at (incmid) {include};
        \draw [->] (vote) -- (threshold) node [midway, left] {Class Confidence};
        \draw [->,dashed] (clause.south) |- ($(clause.south)!.8!(vote.north)$) -| (ta) node [midway, below] {};
        \draw [->,dashed] (vote.south) |- ++(0.5, -0.5) -| (ta)
		node [block, draw=none, text width=, near end, right, fill=white] (fb) {Feedback};
        \draw [->] (threshold) -- ++(1, 0) node [right] {Classification};

        \node [text width=, above=of incmid] (input) {Input Features};
        \draw [->] (input) -- (clause);
        \draw [->] (input) -- (ta);
        \draw [<-] ($(ta.north west)!.75!(ta.north east)$) -- ++(0, 1)
                node [block, text width=, anchor=south, fill=white] (prbg) {Randomizer};

        \node [above=0.5 of prbg] {Learning};

	\path (clause.south) -- (vote.north) node [midway] (clause-res) {};
	\draw [<-, fault] ([yshift=-1em]incmid) -- ++(0,-6mm) node [below, fault block, align=center] {Fault\\Injection};

        \begin{pgfonlayer}{background}
                \draw [draw=none, fill=black!30, rounded corners, outer sep=2mm]
                        ([yshift=1mm]prbg.north west) -| ([shift={(1mm, -1mm)}]fb.south east) -|
                        ([shift={(-1mm, 1mm)}]ta.north west) -- cycle;
        \end{pgfonlayer}
\end{tikzpicture}
        \caption{%
                Overview of the \acf{TM} architecture, with \ac{TA} fault injection highlighted later discussed in \cref{sec:arch-tm-fault-injection}.
        }\label{fig:tm-overview}
\end{figure}
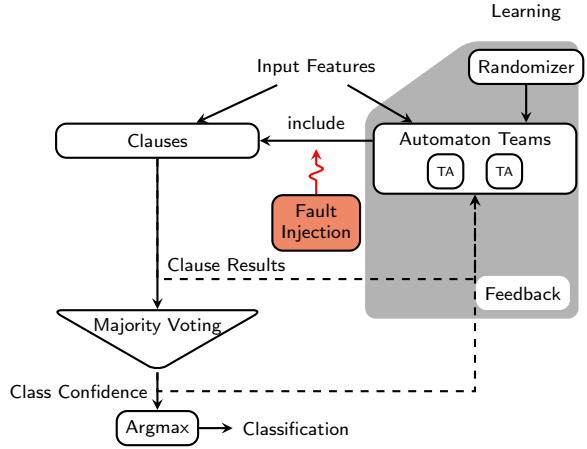

The \ac{TM} is \iac{ML} algorithm which
learns patterns from data features in Boolean form rather than binary numbers. This allows for the algorithm to process data through propositional logic. The main inference component of the \ac{TM} is the
\emph{clause} which composes an \gate*{and} expression of the input Boolean features and
their complements. The \ac{TM} comprises many clauses, each producing a vote.
The composition of each clause is controlled by a vector of \emph{include} bits
(see \cref{fig:tm-overview}). These bits are parameters that are learnt
by teams of \acp{TA}.

Each clause can produce a vote for its class. Half of the clauses can vote
positively, while the other half of the clauses can vote negatively. The
inclusion of inhibition in the voting system enables non-linearity in the
inference process. A majority vote gives an indication of class confidence, which is 
used to classify the input data and influence future decisions of
the automata through the feedback mechanism~\cite{granmo2018tsetlin}.

The \ac{TA} is a class of finite reinforcement automaton~\cite{narendra1989learning}. It produces an
\emph{exclude} output for states below the midpoint, and \emph{include} for
states above the midpoint. The \ac{TA}
receives a penalty or reward from the feedback mechanism based on the current
state of the \ac{TM}. Continued rewards in the end states cause the \ac{TA} to
saturate. A penalty in one of the midstates ($n$ or $n+1$) causes the \ac{TA} to
transition across the decision boundary --- between exclude and
include.

The \ac{TM} controls \ac{TA} feedback through the hyperparameters $T$ and $s$.
$T$ can be thought of as a target for the number of clauses to activate.
Hyperparameter $s$ introduces an element of stochasticity, and therefore exploration, into the \ac{TA}'s behaviour. More details on \ac{TM} can be found in~\cite{granmo2018tsetlin}.


    
\section{FPGA Accelerator Architecture}\label{sec:accel}

\begin{figure}[htbp]
        \centering
        \includegraphics[width=\linewidth]{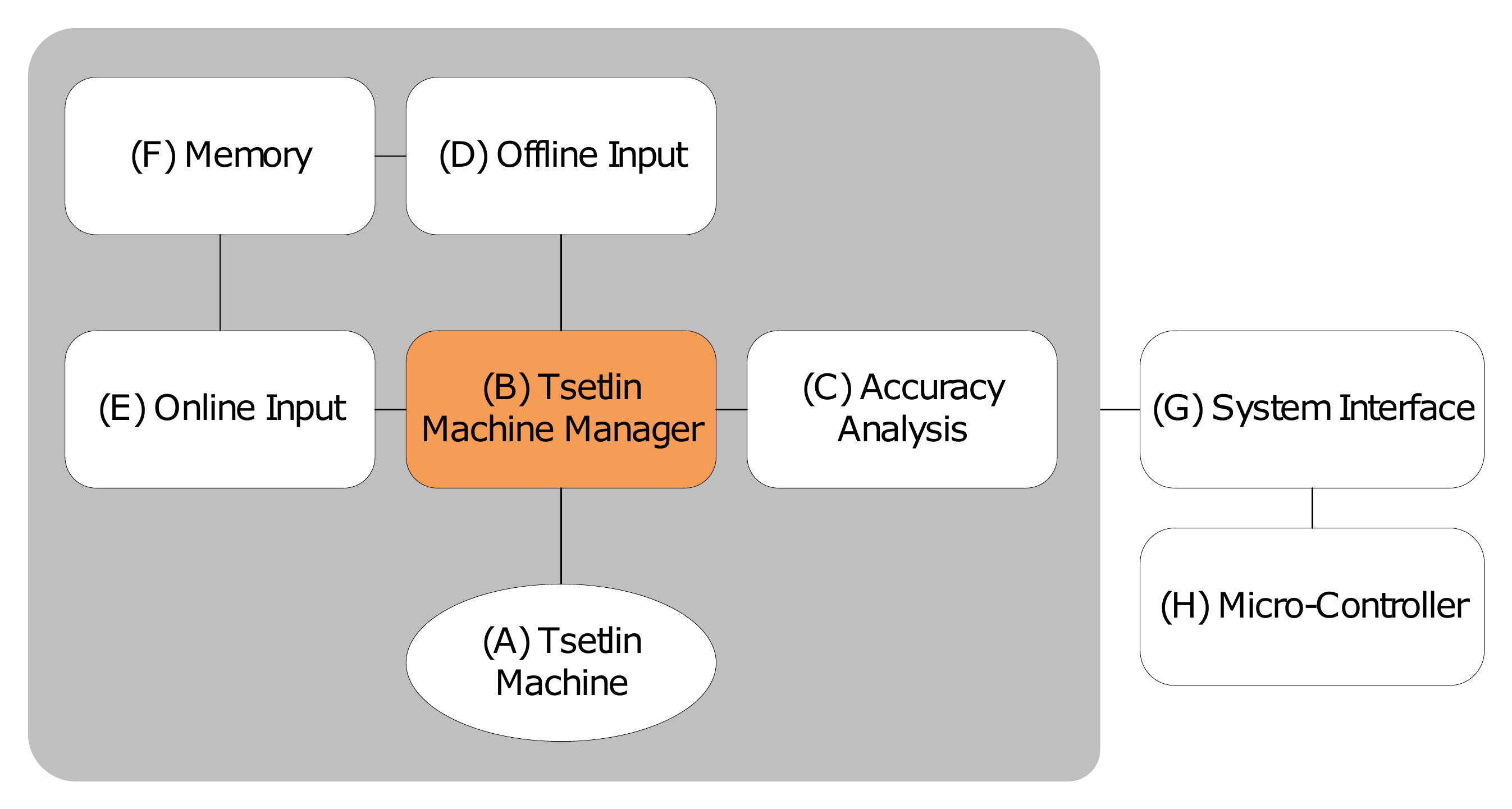}
        \caption{%
                Overview of the proposed online learning architecture.
        }\label{fig:sys-arch-high-level}
\end{figure}

We present a modular \ac{TM} architecture targeted at \acp{FPGA} (\cref{fig:sys-arch-high-level}). The architecture as implemented is capable of both offline and online training to satisfy our need for low energy and at-speed learning as discussed in \cref{sec:intro}. Below we describe the building blocks of the architecture in details.

    \subsection{Tsetlin Machine}
        The \ac{TM} forms the core of the architecture.
        The \ac{TM} hyperparameters such as number of classes, number of clauses and number of \ac{TA} states are parameterized at design time to allow arbitrarily-sized machines to be synthesized. The sensitivity and threshold hyperparameters, $s$ and $T$, are controllable during runtime via \ac{IO} ports. We added two additional features to the \ac{TM} to improve development.
        
        \subsubsection{Over-Provisioning Resources}
            To reduce the need for re-synthesis and to allow adaptation on-the-fly, resources can be over-provisioned. Classes are controlled by a pre-synthesis parameter and can be over-provisioned by not introducing the \ac{TM} to additional classifications during training, which can be utilised to introduce new classifications during online operation.
            Clauses are controlled by a maximum clause number pre-synthesis parameter and a clause number port. They can be over-provisioned by using a number of clauses below the maximum at the clause number port, which can be increased to the maximum during runtime if required. The extent of over-provisioning is application specific.
            
        \subsubsection{Fault Injection}\label{sec:arch-tm-fault-injection}
            Faults present a major issue in systems, therefore we created a way to allow faults to be injected into the system to aid development of fault-tolerant systems~\cite{mathew2014energy}.
            In order to simulate faults in our system we add extra logic to the \ac{TM} which forces the \ac{TA} outputs.
            Each \ac{TA}'s output action is a $1$ or $0$.
            In order to simulate stuck-at faults we introduce \gate*{and}\slash \gates{or} to the output.
            If the ANDed signal is $0$, the output is always $0$ and if the ORed signal is $1$, the \ac{TA} output is always $1$. For fault-free \ac{TA} operation, the ANDed signal should be $1$ and the ORed signal $0$.
            
            A fault controller module provides two mappings for these signals individual to each \ac{TA}.
            These mappings are initially set to $1$ for AND and $0$ for OR, and can then be updated as required.
            Each \ac{TA} is addressable and connected to the microcontroller interface IP, providing access to these mappings from the microcontroller code. This allows various fault configurations to be injected without re-synthesis of the \ac{FPGA} logic.
        
    \subsection{Tsetlin Machine Management}
        The operation of the system is controlled by two main state machines, one for high-level system operations and one for low-level, per data-point, state machine. The high-level manager controls operation of the system as a whole, whereas the low-level manager controls the I/O and operation of the \ac{TM} itself.
            
    \subsection{Accuracy Analysis}
        The accuracy analysis block records the number of errors and total epochs per accuracy analysis cycle. An additional block records the history of these values during simulation in \ac{RAM}, whereas these values can be immediately offloaded to the microcontroller when implemented on an \ac{FPGA} to reduce \ac{RAM} usage.
        
    \subsection{Offline Data Input}
        \subsubsection{Class Filtering}\label{sec:arch-class-filtering}
            In \cref{sec:use-cases-new-class} we investigate the advantages of online training when introducing a new class. Therefore, a filtering subsystem was created, controlled by an external enable signal, to remove a certain class if desired.
        
        \subsubsection{Memory Management}
            The \ac{TM} management subsystem can request data from different sources with data request signals, with the details of these sources abstracted to different modules. Therefore, we created a memory management subsystem to retrieve and parse the required offline data from onboard memory and present it to the \ac{TM} management when required, abstracting the memory interface itself away from the management subsystem.
    
    \subsection{Online Data Input}
        The online data source of a system is application dependent as it may come from a microcontroller or other IP and sensors. Therefore the online data input subsystem was abstracted into multiple layers. This also included the filter IP discussed for the Offline Data Input subsystem.

        \subsubsection{Online Data Manager}
            The \ac{TM} management subsystem requests each row of online data when it requires it. The online data manager provides this data from the online data source via a cyclic buffer.

        \subsubsection{Online Input Data Buffer}
            To allow the \ac{TM} management to be able to periodically check model accuracy, we implemented a cyclic buffer to temporarily store online data in \ac{RAM} to prevent datapoints being ignored by the system during accuracy analysis processes.
        
        \subsubsection{Online Input Parser}
            For experimentation, online data was stored in on-chip ROM, accessed via a cross-correlation interface. We created specific IP to retrieve and parse each row of data, providing it to the system's aforementioned buffer system. Due to this abstraction, this IP can be replaced depending on the online input source without the need to modify the other subsystems.

    \subsection{Memory}
        \subsubsection{Block Memory Manager - Cross-Validation}
            The pattern recognition and inference capabilities of a machine learning model are important. A common way of testing this is to use cross-correlation, where the full dataset is split into subsets for training and validation. These subsets will be referred to as the three data \emph{sets}.
            
            To accommodate both offline and online training, in this system sets are required for offline training, validation and online training. For example, the iris dataset used to acquire our results has 150 unique datapoints. These sets were allocated 30, 60, 60 respectively.
            
            When training \ac{ML} models, data biases can have a significant effect on training~\cite{gu2019understanding}. For example, a specific row of data may result in the \ac{TM} incorrectly classifying a future row of data that would have been classified correctly if the former row had not been used. Additionally, there may be uneven distributions of classes and patterns across these three sets.
            
            Therefore, we took action to mitigate these effects to produce results actually representative of how the system may behave in different scenarios.
            
            We split the full dataset into small subsets. These will be referred to as \emph{blocks}. The length of each block needed to be a factor of each set size, for the iris example the highest common factor, $30$, was chosen. This resulted in $150/30=5$ blocks which could be combined as required to form the larger sets.
            
            The experimentation was re-run for various orderings of these blocks, with the results averaged. The maximum number of possible orderings in this example $5!=120$, so we created a subsystem that could be provided with a set of starting orderings which could then be easily manipulated to produce the full number of orderings.
            
        \subsubsection{Onboard ROM}
            Each block was stored in a separate block ROM, mapped to via the cross-validation IP. Each block ROM was dual port to allow the Online Training set to be used in online training as well as accuracy analysis.
    
    \subsection{System/Microcontroller Interface}
        We created a general piece of IP to provide the on-board microcontrolelr with access so a set of 32-bit I/O registers via an AXI bus. These registers were connected via signals to more specific IP to separate and combine signals into these registers and connect them to the required ports within the system.
    
        Additionally, we implemented a handshaking itnerface between the microcontroller and this more spcific IP. The IP sends a signal to the microcontroller informing it that certain registers are ready to be read from, then pauses the system whilst waiting for the microcontroller to respond. The microcontroller can then read the registers and assert a different signal to convey that the registers have been read successfully. The system detects this signal and resumes operation. This allows the system to operate at high speed without worrying about the microcontroller's speed of operation and race conditions.
    
    \subsection{Microcontroller}
        To reduce development times by reducing the need for re-synthesis and implementation, we made various system parameters controllable and adjustable via a connected microcontroller. Additionally, this allows dynamic control of the system, for example the ability to reconfigure over-provisioned resources.
        The microcontroller also provides a medium to retrieve accuracy information by communicating with a host PC via a connected UART interface and provides an established starting point for acquiring online data via the microcontroller interface. This could provide access to data from sources connected via UART or Ethernet, for example.
    
    \section{System Operation}
\begin{figure}
        \centering
        \includegraphics[width=\linewidth]{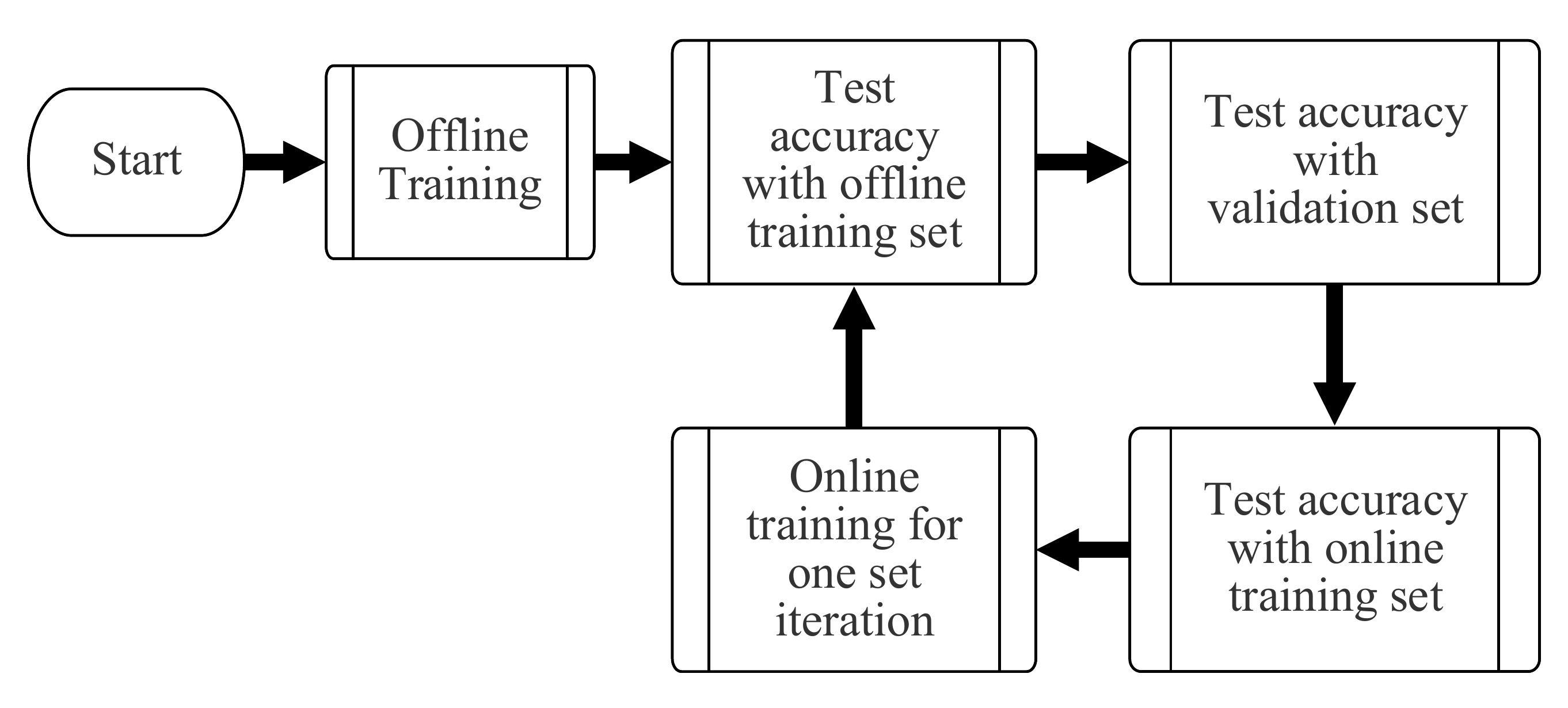}
        \caption{%
                System execution flow.
        }\label{fig:sys-execution-flow}
\end{figure}

The high-level \ac{TM} manager controls execution flow. After initial training using the offline training set, the accuracy is analyzed in terms of the training data and, optionally: a validation set (for inference validation on unseen data) and the online training set.
Online learning is then executed for a set number of datapoints before accuracy analysis is re-run, as shown in \cref{fig:sys-execution-flow}.
This information can be used to determine the effectiveness of online learning over time and with different configurations, as well as to detect possible faults that have occurred.
In offline training and accuracy accuracy analysis for all sets, data is taken from the Offline Input subsystem. During online training, data is taken from the Online Input subsystem.
    
    \section{Experimental Results and Use Cases}\label{sec:use-cases}
The effectiveness of the system was investigated and demonstrated through three possible use cases. For all experimentation, the \emph{iris} dataset was used ($16$ booleanised inputs, $3$ classifications, $150$ unique datapoints) with the same system parameters: $16$ clauses, $s$ hyper-parameter $1.375$ for offline and $1$ for online, $T$ hyper-parameter $15$, $10$ offline training iterations and 120 cross-correlated orderings with the results averaged. These values were carefully chosen from offline experiments as they demonstrated good learning efficacy for the particular dataset. An additional promising application of this system, with its cross-validation infrastructure, is rapid hyper-parameter search. The fast execution time allows entire datasets to be analyzed in a matter of seconds, allowing the optimum hyper-parameters for a given dataset to be discovered within a short period of time. 

\subsection{Limited Initial Training Data}
    In many applications, the number of initial data points may be limited. Online learning provides further training during operation.
    
    We investigated the effect of online learning using labelled data. Often it may be possible to acquire some labelled data at runtime, for example from user feedback or to an \ac{IoT} edge node from a central system. Alternatively, it may be possible to use a confident estimated classification as the class for a set of inputs.
    
\begin{figure}
        \centering
        \includegraphics[scale=\scaleplot]{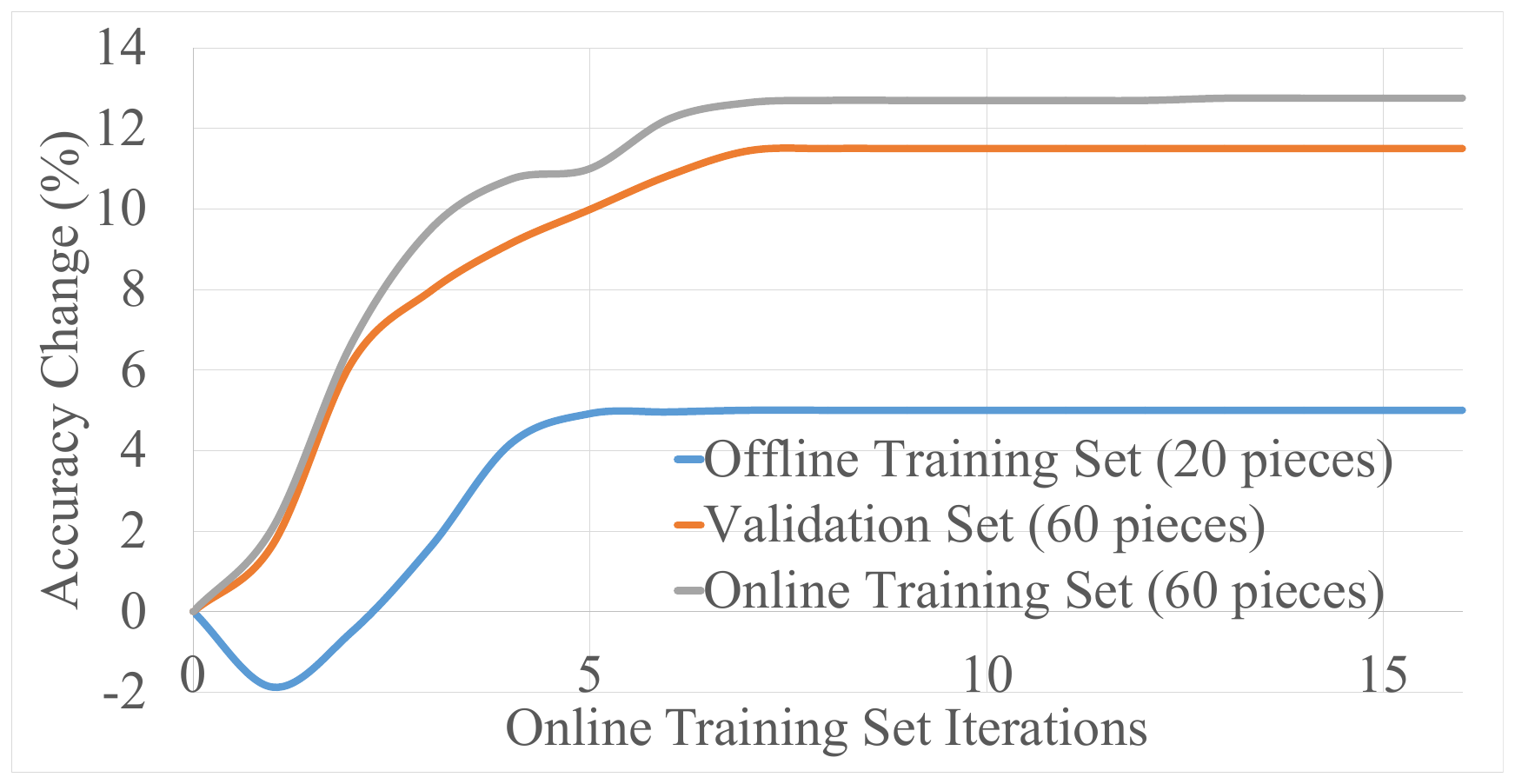}
        \caption{%
                The effect of online learning with labelled data on accuracy. Starting accuracies: offline training set $83\%$, validation set $79.5\%$ and online training set $79.5\%$.
        }\label{fig:results-online-labelled}
    \end{figure}

    An offline training set of length $20$ was initially used to train the \ac{TM} over $10$ iterations. The accuracy was then analyzed for this offline training set, a validation set and an online training set.
    Online training with $60$ labelled datapoints was carried out for a single iteration of the set, with a small $s$ hyper-parameter value of $1$. A lower $s$ value increases the likelihood of inaction, so overall there will be a bias away from issuing feedback when a low $s$ value is used, resulting in reduced power consumption. The accuracy was re-analyzed and the process repeated for 16 iterations of the online training set. \cref{fig:results-online-labelled} shows the resulting effect on accuracy due to this online training. It shows the accuracy of the validation and online training sets increased similarly by approximately $12\%$, unlike the offline training set accuracy which increased less significantly by only approximately $5\%$.
    
    The accuracy of the online training set increased the most as the training was being performed using this set, so the feedback applied to the \ac{TA}s was targeted at these particular datapoints. The patterns contained within the full dataset were spread across the three sets. Therefore, training on the online training set resulted in an increase in accuracy for the validation set due to these shared patterns being exposed to the \ac{TM} during online training.
    As the offline training set was used for initial offline training, the \ac{TA}s were initially configured for the offline training set datapoints meaning the offline training set had the largest initial accuracy. When training with the online training set, the \ac{TA} states changed to also fit the online training set. The accuracy for the offline training set ultimately increased, again due to patterns being shared across the three sets, but not as significantly as the other two.
    This was primarily due to the higher initial accuracy, but also links to catastrophic forgetting where the model may lose its fit for old data as new data is learned~\cite{delange2021continual}. It could be advantageous to use a replay method, continuing training with occasional datapoints from the offline training set during online operation~\cite{delange2021continual}.

\subsection{Unseen Class Introduction}\label{sec:use-cases-new-class}
    It is possible for additional classes to be introduced during online operation. Traditionally this support would not be possible as the system would require full retraining, but with online learning the system can adapt dynamically.
    
    For testing, we created the class filter IP, in \cref{sec:arch-class-filtering}, to remove a desired class during offline training and initial online operation, to experiment using online learning in this scenario by introducing a previously unseen classification midway through online operation.
    
    We again used labelled data in the online training set, but further experimentation could involve analyzing class confidence to determine the likelihood a datapoint belongs to an unseen classification and provide appropriate feedback.
    
\begin{figure}
        \centering
        \includegraphics[scale=\scaleplot]{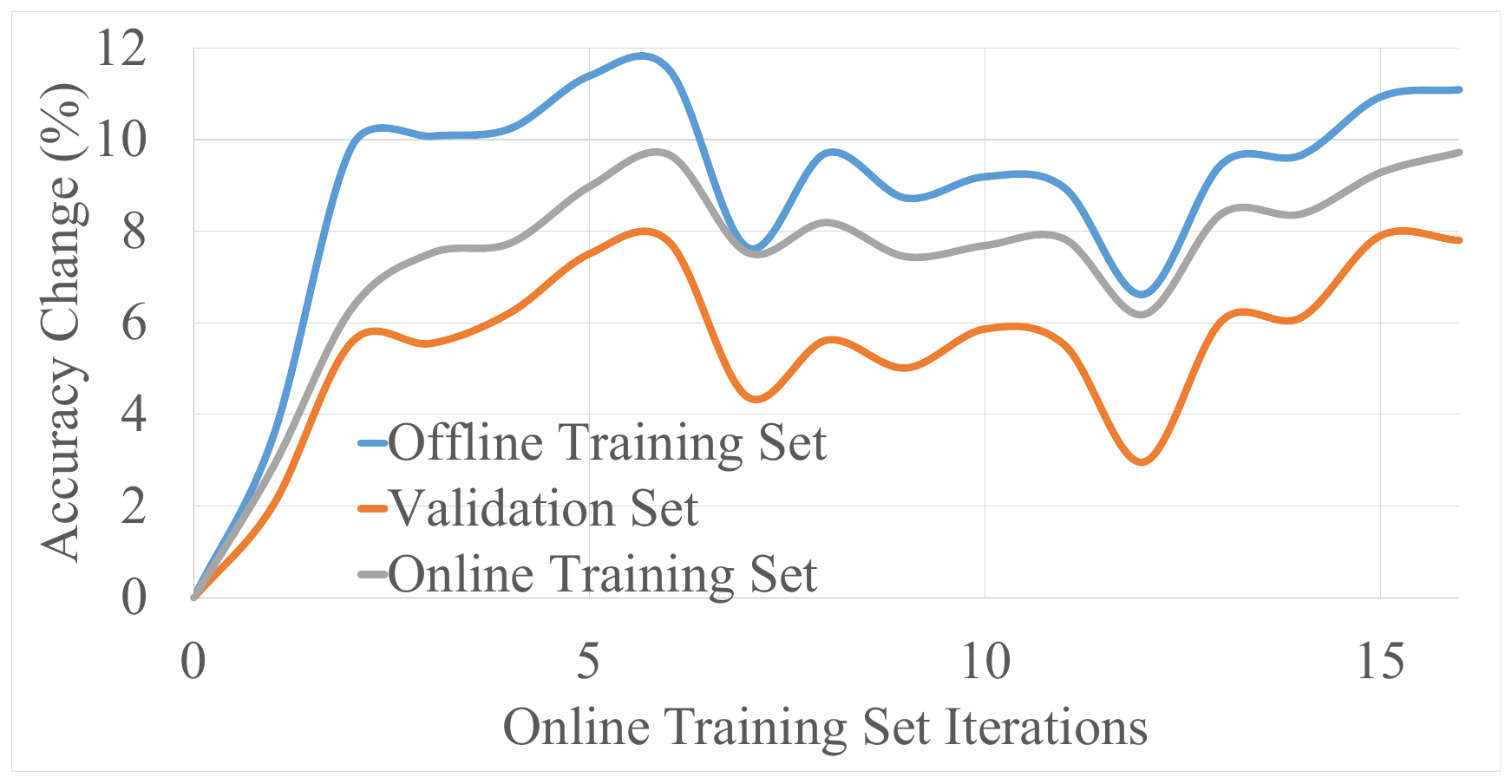}
        \caption{%
                Accuracy results with the \emph{iris} dataset's $0$ class filtered from the three sets for the entire operation. Starting accuracies: offline training set $75.75\%$, validation set $79.66\%$ and online training set $79.66\%$.
        }\label{fig:results-removed-class-online-baseline}
    \end{figure}

    Initially, the accuracy over $16$ online learning iterations was analyzed, but with one classification filtered out. $30$ datapoints were available in the offline training set, but only $20$ were used in previous experimentation. Therefore, the size of the set after filtering one of three classes out was still approximately $20$. The validation and online training sets had no such redundancy, so were each reduced to approximately $40$ in size when one of three datasets was filtered out.
    The results of this are demonstrated in \cref{fig:results-removed-class-online-baseline}, showing an increase in accuracy over online training. Oscillations were present, due to the small nature of the dataset and relatively small number of online iterations, but this provided a suitable baseline. As expected, the validation set increased in accuracy less than the online training set, but the accuracy of the offline training set increased the most significantly. As the initial accuracy was lower for it in this experiment, it allowed the accuracy to increase further during online operation.
    
\begin{figure}
        \centering
        \includegraphics[scale=\scaleplot]{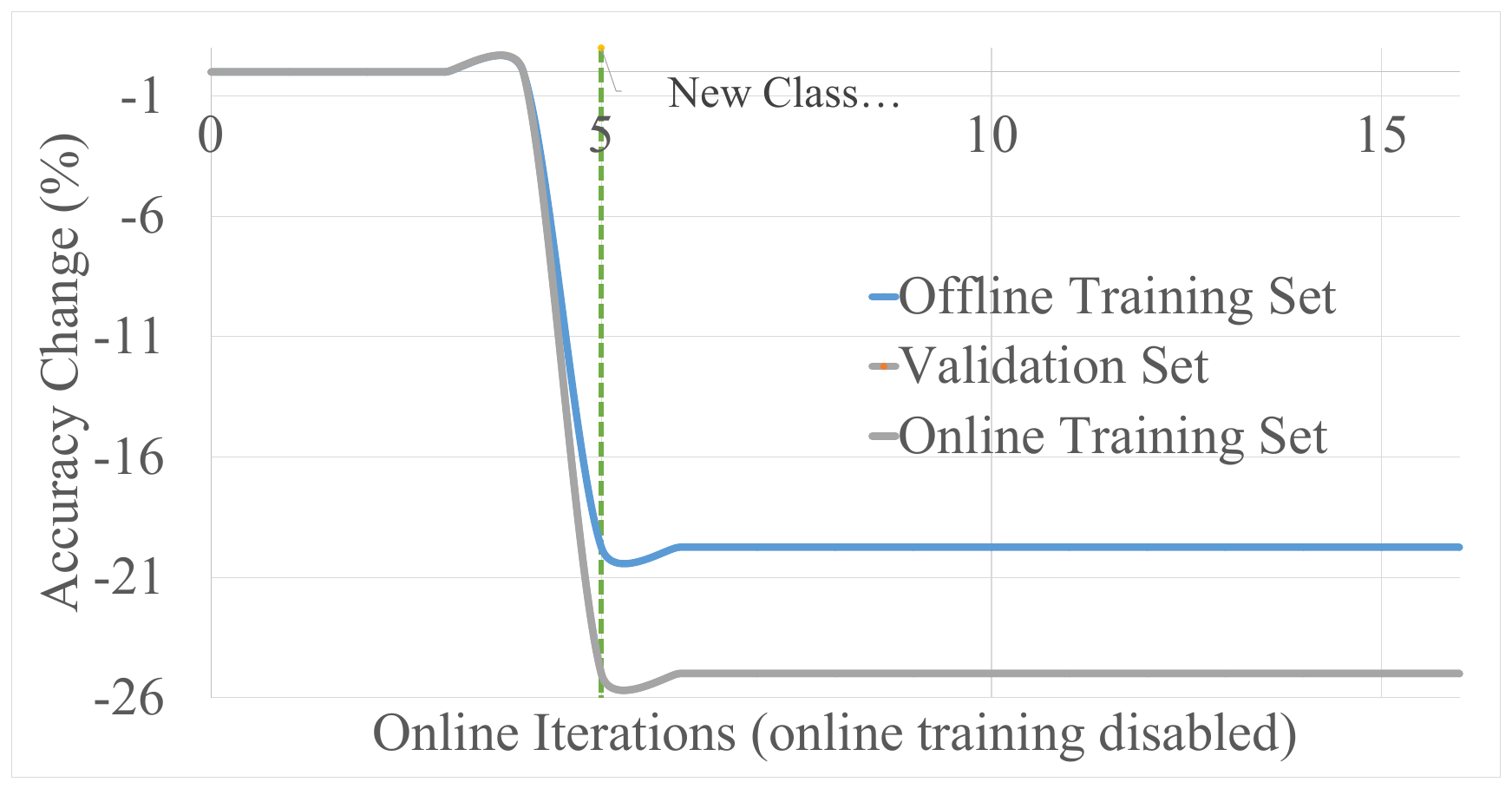}
        \caption{%
                Effect on accuracy of introducing new classification at runtime (after $5$ online iterations) with online training disabled. Starting accuracies: offline training set $75.75\%$, validation set $79.66\%$ and online training set $79.66\%$.
        }\label{fig:results-new-class-baseline}
    \end{figure}

    We repeated this to produce a second baseline with online learning disabled and a new classification introduced after $5$ online iterations, with a decrease in accuracy observed in \cref{fig:results-new-class-baseline} for all sets when the unseen class was introduced. The decrease was the same for the validation and online training sets, but was different for the offline training set with the offline training set being a third the size of the other two contributing to this discrepancy.
    
\begin{figure}
        \centering
        \includegraphics[scale=\scaleplot]{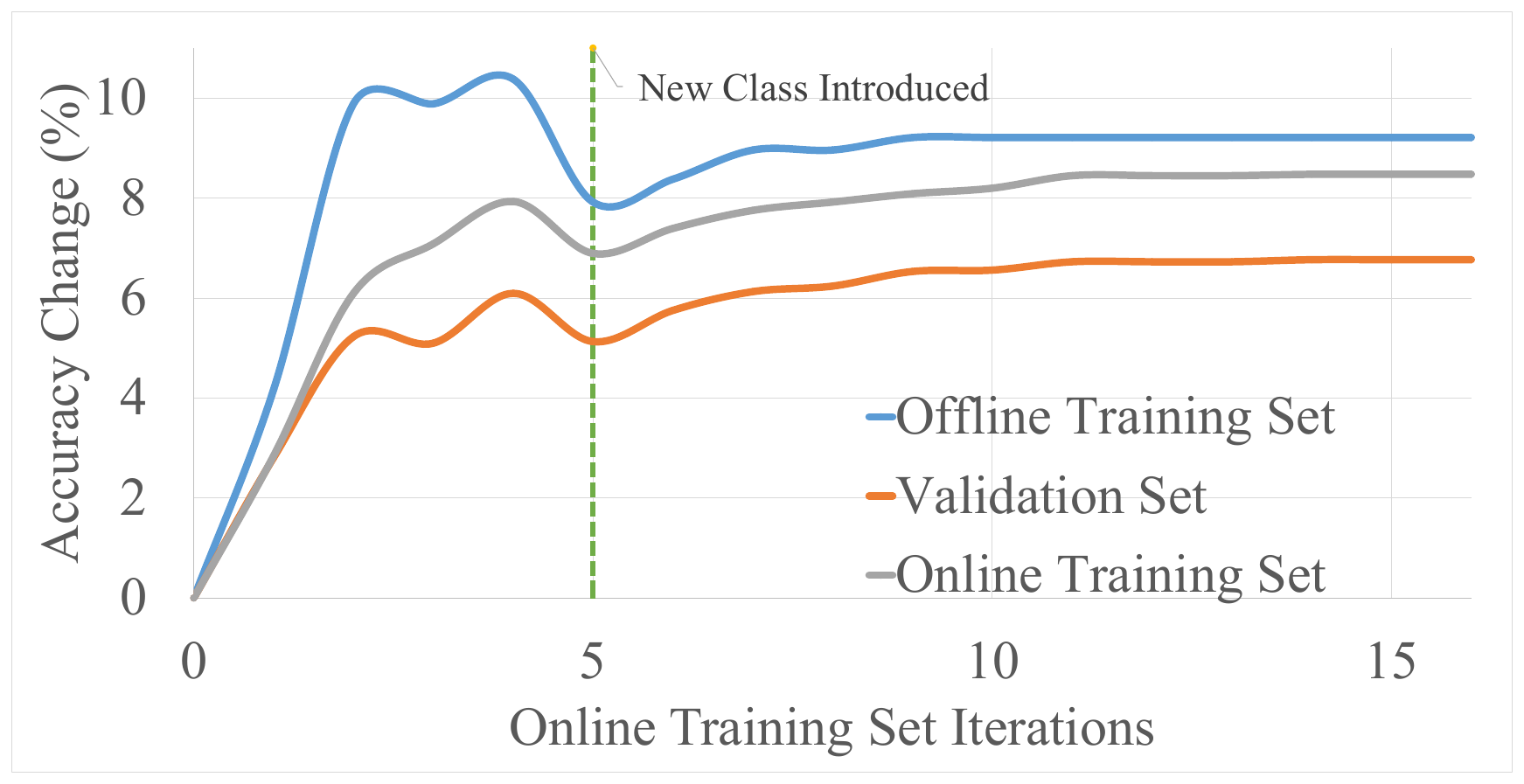}
        \caption{%
                Effect of online training on new classification introduction. Starting accuracies: offline training set $75.75\%$, validation set $79.66\%$ and online training set $79.66\%$.
        }\label{fig:results-new-class-online-enabled}
    \end{figure}

    Finally, we combined these two experiments to observe the effect of labelled online learning on \ac{TM} accuracy with the introduction of a class at runtime, the results being observed in \cref{fig:results-new-class-online-enabled}. The accuracy initially increased inline with the baseline in \cref{fig:results-removed-class-online-baseline} before the new class was introduced. When the new class was introduced after $5$ online iterations, the accuracy briefly decreased (after the new class was introduced, one iteration of online training was executed before the accuracy was re-analyzed) whilst the \ac{TM} began being trained with the online training set containing the new class. The accuracy soon recovered, showing a significantly positive outcome compared to without online training. The relative accuracy increase for these sets were comparable after the point where the new class was introduced.
    
\subsection{Fault Mitigation}
    Faults can be major issues, especially in unsupervised edge nodes. An example of a fault causing an issue within a \ac{TM} is a \ac{TA} getting stuck outputting $0$ or $1$, regardless of the literal input to the \ac{TA}. A \ac{TM} relies on stochasticity and majority voting, therefore there is more than one training solution. So, if such a fault exists before training, the system can train itself in a way that this \ac{TA} has little effect on the output. Therefore when online learning is used, training continues after initial training, having the potential to re-train ``around" faulty \ac{TA}s.
    Faults can be injected into the \ac{TM} and the effects observed with and without online learning enabled. Usually the number of faults is small, but for demonstration purposes a large number of faults were injected at the same time.
    
    \subsubsection{Runtime fault mitigation using labelled online learning}
        A Python script was created and used to create an equal spread of fault mappings across the \ac{TA}s.
        The \ac{TM} is effective at fault mitigation when training \cite{shafik2020explainability}, so to observe the effects of faults on a fully-trained system the faults were injected after $5$ online iterations.
        
\begin{figure}
            \centering
            \includegraphics[scale=\scaleplot]{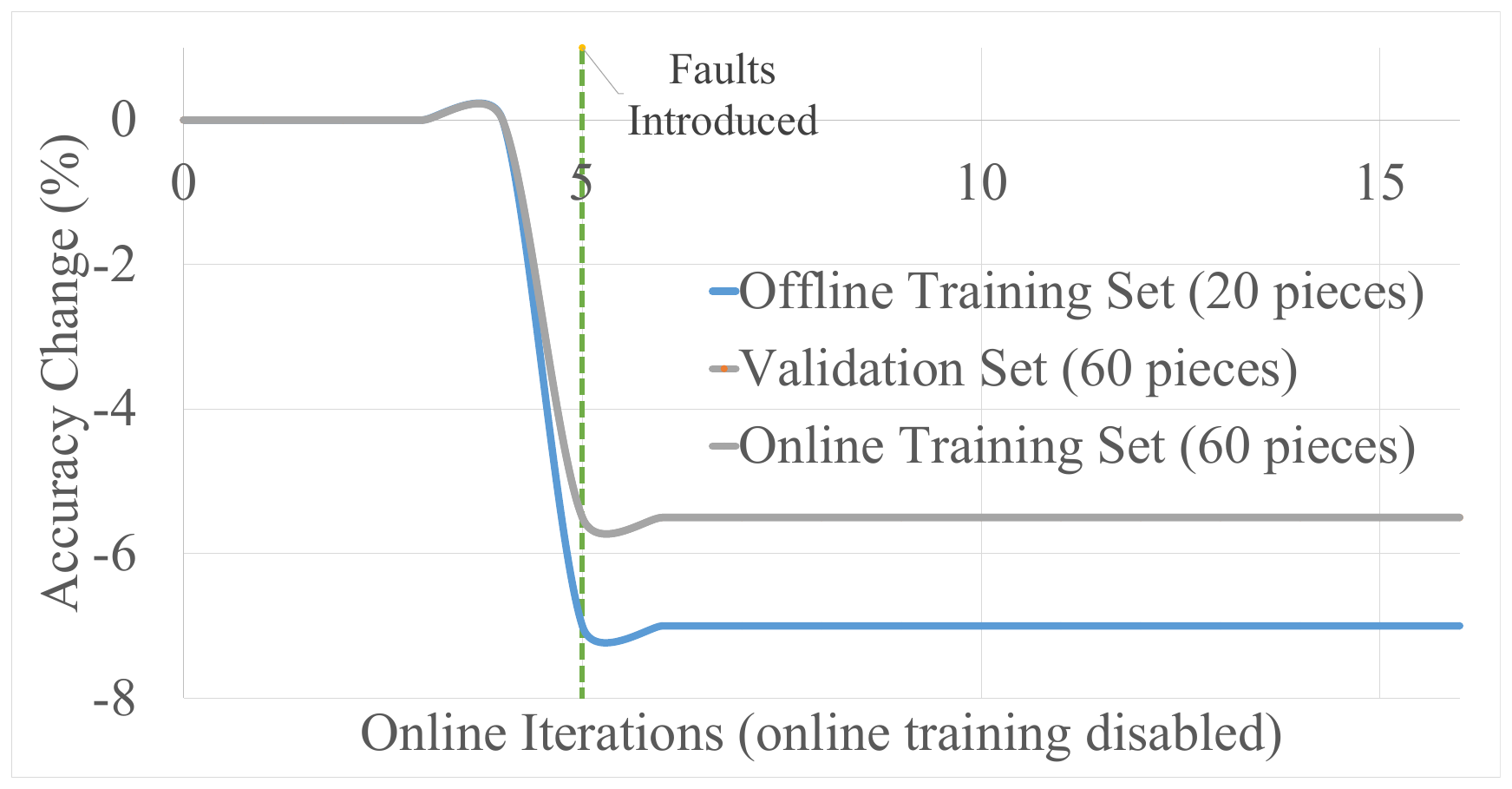}
            \caption{%
                    Faults introduced after 5 online iterations, with online learning disabled. 20\% stuck at 0 faults evenly distributed across \ac{TA}s. Starting accuracies: offline training set $83\%$, validation set $79.5\%$ and online training set $79.5\%$.
            }\label{fig:results-faults-baseline}
        \end{figure}

        The faults injected consisted of forcing the output of $20\%$ of the \ac{TA}s to $0$. The effect with online training disabled was observed in \cref{fig:results-faults-baseline}, where the accuracy decreased as expected. The difference in accuracy change between the offline training set and the validation/online training sets was due to the difference in set size.
        
\begin{figure}
            \centering
            \includegraphics[scale=\scaleplot]{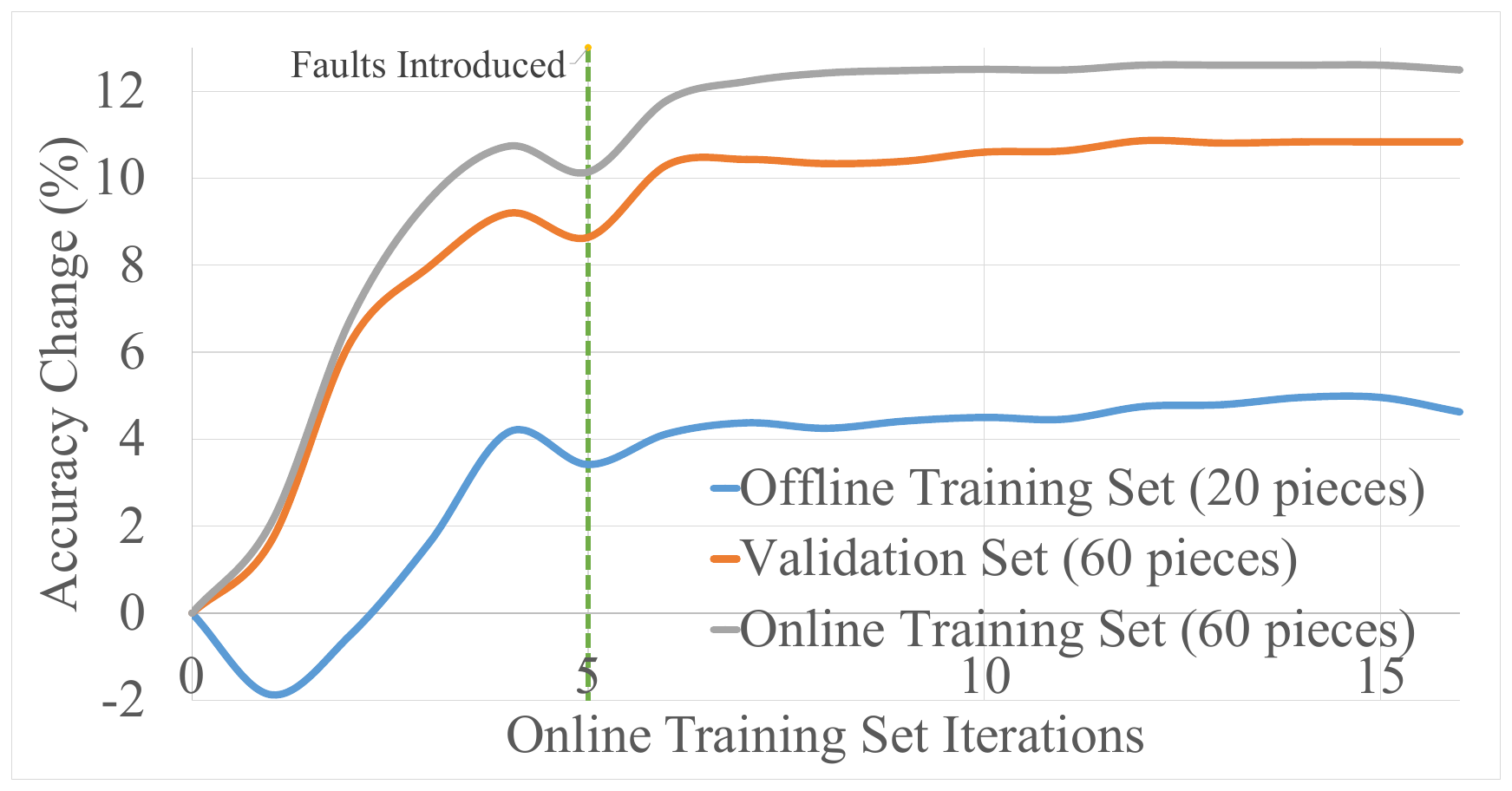}
            \caption{%
                    Faults introduced after 5 online iterations, with online learning enabled. 20\% stuck at 0 faults evenly distributed across \ac{TA}s. Starting accuracies: offline training set $83\%$, validation set $79.5\%$ and online training set $79.5\%$.
            }\label{fig:results-faults-online-enabled}
        \end{figure}

        This was repeated, but with online learning enabled. \cref{fig:results-faults-online-enabled} demonstrated that the accuracy briefly fell (after the faults were injected, one iteration of online training was executed before the accuracy was re-analyzed) before starting to recover with further online iterations, with the final accuracy increases after $16$ iterations being on par with the fault-free system demonstrated in \cref{fig:results-online-labelled}. After faults were injected, the relative changes in accuracy across the three sets were comparable.
        This was significantly better than the offline learning demonstration in \cref{fig:results-faults-baseline}, due to the stochasticity and majority voting properties of the \ac{TM}: feedback is applied individually for each \ac{TA}, so issues with other \ac{TA}s may result in feedback being applied differently to non-faulty \ac{TA}s that it would have been otherwise, leading to clause characteristics changing. For example, a clause voting against a classification may change to cancel out a clause containing faults causing it to always vote for a classification.
        
    \subsubsection{Further Mitigation Strategies}
        After every set number of online learning epochs, the \ac{TM} accuracy is analyzed using the initial training data as well as unseen validation data. This accuracy analysis can be used to enable/disable online learning, control online learning sensitivity and to choose to fully retrain the \ac{TM} on-chip if the accuracy has fallen below a certain threshold (i.e. significant faults have occurred). Additionally, with over-provisioning of clauses, additional clauses can be enabled for this retraining to further mitigate the effect of faulty \ac{TA}s \cite{shafik2020explainability}.
    
    \section{Performance and Power Consumption}
The parallel nature of a hardware-implemented \ac{TM} is unrivalled by software implementations due to the high degree of concurrency. Where a software implementation cannot calculate all clause outputs at the same time, our hardware \ac{TM} implementation takes just two clock cycles to complete inference and feedback for all clauses and \ac{TA}s. The throughput of the \ac{TM} is one datapoint per clock, requiring an additional clock cycle to buffer the I/O. This decreases execution times from minutes, or longer, on a computer using a software implementation down to a matter of seconds~\cite{wheeldon2020learning}. Whilst the system can interface with a microcontroller, it contains handshake logic for this interaction to allow the speed of microcontroller operation to be independent from the speed of the hardware \ac{TM}. The only possible slowdown is stalling during this handshaking process.

When inference or learning is not occurring, the \ac{TM} is clock-gated, providing significant power consumption improvements. Additionally, clock gating is used for the over-provisioned clauses and \ac{TA}s to reduce the power consumption overhead when they remain unused. The estimated power consumption for the experimental implementation on a Xilinx Zybo Z7-20 was 1.725W, of which 1.4W was associated with the onboard microcontroller (using default tool activity).
    
    \section{Conclusions}
We demonstrated a new \ac{TM} based online learning hardware architecture, implemented on \ac{FPGA}s, leveraging their flexibility to allow rapid development. The architecture provides a fast and reliable way of utilizing on-chip learning in both offline and online circumstances. Online learning was demonstrated with labelled data, demonstrating its positive effects on accuracy in three use cases. Three different use cases were presented in order to validate the effectiveness and flexibility of the proposed architecture. We showed that the architecture is capable of at-speed and low-power online learning using a flower type detection dataset. 

Additional research directions will involve experimentation with the \ac{TM}'s classification confidence to apply feedback when using unlabelled online data, as well as using the class confidences from each class to determine if unlabelled data may belong to an unseen classification. Additionally, it may be investigated how continuous accuracy analysis (every $N$ cycles test the accuracy with a single piece of offline training data, maintaining a cumulative average) can be used to detect faults and trigger system retraining/resource re-provisioning and the impact of injecting faults at the clause output level.

    \bibliographystyle{abbrv}
    \bibliography{adrian}

\begin{thebibliography}{10}

\bibitem{bhattarai2020measuring}
B.~Bhattarai, O.-C. Granmo, and L.~Jiao.
\newblock Measuring the {{Novelty}} of {{Natural Language Text Using}} the
  {{Conjunctive Clauses}} of a {{Tsetlin Machine Text Classifier}}.
\newblock {\em arXiv:2011.08755 [cs]}, Nov. 2020.

\bibitem{biswas2017machine}
D.~Biswas, V.~Balagopal, R.~Shafik, B.~M. Al-Hashimi, and G.~V. Merrett.
\newblock Machine learning for run-time energy optimisation in many-core
  systems.
\newblock In {\em Design, Automation \& Test in Europe Conference \& Exhibition
  (DATE), 2017}, pages 1588--1592. IEEE, 2017.

\bibitem{delange2021continual}
M.~Delange, R.~Aljundi, M.~Masana, S.~Parisot, X.~Jia, A.~Leonardis,
  G.~Slabaugh, and T.~Tuytelaars.
\newblock A continual learning survey: {{Defying}} forgetting in classification
  tasks.
\newblock {\em IEEE Trans. Pattern Anal. Mach. Intell.}, pages 1--1, 2021.

\bibitem{fontenla2013online}
{\'O}.~Fontenla-Romero, B.~Guijarro-Berdi{\~n}as, D.~Martinez-Rego,
  B.~P{\'e}rez-S{\'a}nchez, and D.~Peteiro-Barral.
\newblock Online machine learning.
\newblock In {\em Efficiency and Scalability Methods for Computational
  Intellect}, pages 27--54. IGI Global, 2013.

\bibitem{granmo2018tsetlin}
O.-C. Granmo.
\newblock The {{Tsetlin Machine}} -- {{A Game Theoretic Bandit Driven
  Approach}} to {{Optimal Pattern Recognition}} with {{Propositional Logic}}.
\newblock {\em arXiv:1804.01508 [cs]}, Apr. 2018.

\bibitem{gu2019understanding}
J.~Gu and D.~Oelke.
\newblock Understanding bias in machine learning.
\newblock {\em arXiv preprint arXiv:1909.01866}, 2019.

\bibitem{lei2020arithmetic}
J.~Lei, A.~Wheeldon, R.~Shafik, A.~Yakovlev, and O.-C. Granmo.
\newblock From arithmetic to logic based {{AI}}: {{A}} comparative analysis of
  neural networks and tsetlin machine.
\newblock In {\em 2020 27th {{IEEE}} Int. {{Conf}}. {{Electron}}. {{Circuits}}
  Syst.}, 2020.

\bibitem{london1999empowered}
M.~London and J.~W. Smither.
\newblock Empowered self-development and continuous learning.
\newblock {\em Human Resource Management: Published in Cooperation with the
  School of Business Administration, The University of Michigan and in alliance
  with the Society of Human Resources Management}, 38(1):3--15, 1999.

\bibitem{maheshwari2023redress}
S.~Maheshwari, T.~Rahman, A.~Yakovlev, A.~Rafiev, L.~Jiao, O.-C. Granmo, et~al.
\newblock Redress: Generating compressed models for edge inference using
  tsetlin machines.
\newblock {\em IEEE Transactions on Pattern Analysis and Machine Intelligence},
  2023.

\bibitem{mathew2014energy}
J.~Mathew, R.~Shafik, and D.~Pradhan.
\newblock {\em Energy-efficient fault-tolerant systems}.
\newblock Springer USA, 2014.

\bibitem{napolitano1999complete}
M.~R. Napolitano, G.~Molinaro, M.~Innocenti, B.~Seanor, and D.~Martinelli.
\newblock A complete hardware package for a fault tolerant flight control
  system using online learning neural networks.
\newblock In {\em Proceedings of the 1999 American Control Conference (Cat. No.
  99CH36251)}, volume~4, pages 2615--2619. IEEE, 1999.

\bibitem{narendra1989learning}
K.~S. Narendra and M.~A.~L. Thathachar.
\newblock {\em Learning Automata: An Introduction}.
\newblock Prentice-Hall, Inc., USA, 1989.

\bibitem{saha2021relational}
R.~Saha, O.-C. Granmo, V.~I. Zadorozhny, and M.~Goodwin.
\newblock A {{Relational Tsetlin Machine}} with {{Applications}} to {{Natural
  Language Understanding}}.
\newblock {\em arXiv:2102.10952 [cs]}, Feb. 2021.

\bibitem{sahoo2017online}
D.~Sahoo, Q.~Pham, J.~Lu, and S.~C. Hoi.
\newblock Online deep learning: Learning deep neural networks on the fly.
\newblock {\em arXiv preprint arXiv:1711.03705}, 2017.

\bibitem{shafik2020explainability}
R.~Shafik, A.~Wheeldon, and A.~Yakovlev.
\newblock Explainability and dependability analysis of learning automata based
  {{AI}} hardware.
\newblock In {\em 2020 {{IEEE}} 26th Int. {{Symp}}. {{On}}-Line Test.
  {{Robust}} Syst. {{Des}}.}, pages 1--4. {IEEE}, July 2020.

\bibitem{wang2021enabling}
E.~Wang, J.~J. Davis, D.~Moro, P.~Zielinski, C.~Coelho, S.~Chatterjee, P.~Y.~K.
  Cheung, and G.~A. Constantinides.
\newblock Enabling binary neural network training on the edge, 2021.

\bibitem{wheeldon2020learning}
A.~Wheeldon, R.~Shafik, T.~Rahman, J.~Lei, A.~Yakovlev, and O.~C. Granmo.
\newblock Learning automata based energy-efficient {{AI}} hardware design for
  {{IoT}} applications.
\newblock {\em Philos Trans R Soc Math Phys Eng Sci}, 378(2182), Oct. 2020.

\bibitem{wheeldon2020lowlatency}
A.~Wheeldon, A.~Yakovlev, R.~Shafik, and J.~Morris.
\newblock Low-{{Latency Asynchronous Logic Design}} for {{Inference}} at the
  {{Edge}}.
\newblock In {\em To Appear in {{Proc DATE}}'21}, Dec. 2020.

\bibitem{yu2021online}
H.~Yu, H.~Xie, X.~Yang, H.~Zou, and S.~Gao.
\newblock Online sequential extreme learning machine with the increased
  classes.
\newblock {\em Computers \& Electrical Engineering}, 90:107008, 2021.

\end{thebibliography}
\end{document}